%% file: main.tex
\documentclass{egpubl}
\usepackage{pg2020}
 
 \SpecialIssuePaper         

\usepackage[T1]{fontenc}
\usepackage{dfadobe}  
\usepackage{cite}  

\BibtexOrBiblatex
\electronicVersion
\PrintedOrElectronic
\ifpdf \usepackage[pdftex]{graphicx} \pdfcompresslevel=9
\else \usepackage[dvips]{graphicx} \fi

\usepackage{egweblnk} 

\usepackage{times}
\usepackage{epsfig}
\usepackage{graphicx}
\usepackage{amsmath}        
\usepackage{amssymb}
\usepackage{bm}
\usepackage{booktabs}
\usepackage{multirow}
\usepackage{balance}
\usepackage{enumitem}
\renewcommand{\vec}{\mathbf}
\DeclareMathOperator*{\argmin}{argmin} 
\newcommand{\etal}{\textit{et al.}}

\title[Monocular Human Pose and Shape Reconstruction using Part Differentiable Rendering]%
      {Monocular Human Pose and Shape Reconstruction using \\ Part Differentiable Rendering}

\author[M. Wang et al.]
{\parbox{\textwidth}{\centering Min Wang$^{1}$\orcid{0000-0002-3946-7439}, Feng Qiu$^{1}$\orcid{0000-0002-3946-7439}, Wentao Liu$^{2}$\orcid{0000-0001-6587-9878}, Chen Qian$^{2}$\orcid{0000-0002-8761-5563}, Xiaowei Zhou$^{3}$\orcid{0000-0003-1926-5597}, and Lizhuang Ma$^{1}$\thanks{Corresponding author: ma-lz@cs.sjtu.edu.cn (Lizhuang Ma)}\orcid{0000-0003-1653-4341}}
        \\
{\parbox{\textwidth}{\centering $^{1}$Shanghai Jiao Tong University, China \\ $^{2}$SenseTime Research, China  \\ $^{3}$Zhejiang University, China
}
}
}

\begin{document}


\maketitle
\begin{abstract}
Superior human pose and shape reconstruction from monocular images depends on removing the ambiguities caused by occlusions and shape variance. Recent works succeed in regression-based methods which estimate parametric models directly through a deep neural network supervised by 3D ground truth. However, 3D ground truth is neither in abundance nor can efficiently be obtained.
In this paper, we introduce body part segmentation as critical supervision. Part segmentation not only indicates the shape of each body part but helps to infer the occlusions among parts as well. To improve the reconstruction with part segmentation, we propose a part-level differentiable renderer that enables part-based models to be supervised by part segmentation in neural networks or optimization loops.
We also introduce a general parametric model engaged in the rendering pipeline as an intermediate representation between skeletons and detailed shapes, which consists of primitive geometries for better interpretability.
The proposed approach combines parameter regression, body model optimization, and detailed model registration altogether. Experimental results demonstrate that the proposed method achieves balanced evaluation on pose and shape, and outperforms the state-of-the-art approaches on Human3.6M, UP-3D and LSP datasets.

\end{abstract}  
\section{Introduction}
\input{fig_motivation.tex}

Human body reconstruction from a single image is a challenging task due to the ambiguities of flexible poses, occlusions, and various somatotypes. The reconstruction task aims at predicting both human pose and shape parameters. A large variety of applications such as motion capture, virtual or augmented reality, human-computer interaction, visual effects, and animations rely on accurately recovered body models.

Recent results have shown that optimizing 3D-2D consistency between a 3D human body model and 2D image cues is beneficial to both fitting and regression-based methods \cite{bogo2016keep,lassner2017unite,Kanazawa_2018_CVPR}.
With the emergence of large annotated datasets\cite{andriluka20142d,lin2014microsoft} and deep learning architectures\cite{newell2016stacked,cao2017realtime,sun2019deep}, we can acquire robust 2D keypoints prediction of the human joints.
Previous methods\cite{bogo2016keep, lassner2017unite} fit the body model primarily with 2D keypoints to achieve consistent 3D results.
However, only 2D keypoints are not sufficient for the alignment between the 3D model and images. We argue that body part segmentation is a compelling image cue for occlusion reasoning. As shown in Figure~\ref{fig:motivation.}(a), the predicted model is consistent with ground truth 2D keypoints, but the pose is still incorrect as two forearms are at wrong places along with depth. In contrast, with the correct part segmentation in Figure~\ref{fig:motivation.}(b), we can align the model as consistent as the real person in the image.

However, how to utilize part segmentation as supervision in learning-based approaches is still an open question. We commonly generate the part segmentation from a 3D model by a rendering pipeline, which contains a non-differentiable process called rasterization. 
The modern rasterizer blocks the back-propagation in the training process, and derivatives cannot be propagated from the image level loss to the vertices of the 3D mesh.
In order to achieve a differentiable renderer, some impressive efforts have been made to compute derivatives correlated to the perspective loss in recent years, such as OpenDR\cite{loper2014opendr}, Neural Mesh Renderer\cite{omran2018neural} and Soft Rasterizer\cite{liu2019soft}. We formulate rasterization in a part-individual way, not only making use of the perspective loss on rendered images but also inferring the correlations of parts on the depth dimension (as known as z-axis). The paradigm of part differentiable rendering distinguishes our approach from existing works. As predicting part segmentation from images becomes more reliable\cite{lin2017refinenet}, the reconstruction results could be more accurate due to fewer ambiguities caused by occlusions.

To reconstruct the pose and shape with part segmentation, representation of the human body should be simple, part-based, parametric and looks roughly like a real body. One of the most widely used models in human pose estimation is the skeletal representation that consists of 3D keypoint locations and kinematic relationships. It is a simple yet effective manner but commonly impractical to reason about occlusion, collision and the somatotype. In contrast, 3D pictorial structures \cite{felzenszwalb2005pictorial} composed of primitive geometries like cylinders, spheres and ellipsoids are adopted frequently in traditional works. This kind of model can be modulated by parameters explicitly, but it does not look like a real person. Detailed human models, like SMPL \cite{loper2015smpl}, lead to a significant success in human shape reconstruction, whose shape prior is learnt from thousands of scanned bodies. Nevertheless, SMPL depends on a template mesh which means we have to retrain the prior when using a new template. Another problem is that the shape parameters of most detailed models are in the latent space, so it is hard to change body parts independently by the shape parameters. 

Our goal is to propose an intermediate representation to combine the advantages of these models. In one hand, the proposed model has a binding on the skeletal representation to benefit pose estimation. In the other hand, the intermediate model, which is composed of primitive geometries, is applicable for shape reconstruction. It could also be embedded in a detailed human mesh so that we can build the mapping function to a detailed model by minimizing the difference between them. In practice, we predict the parameters of the intermediate model from the 2D image to represent a rough shape of the objective body, and then map to any detailed models.

In this paper, we propose a Part-level Differentiable Renderer called \emph{PartDR}. To leverage the part segmentation as supervision, we use this module to compute derivatives of the loss, which indicates the difference between the silhouettes of the rendered and real body parts. Then each vertex obtains the gradient by backward propagation. Compared with the object-level Neural Mesh Renderer (NMR)\cite{kato2018renderer}, we design an occlusion-aware loss in the part renderer to identify the occlusions between different parts and keep occlusion consistency with part segmentation.
In addition, we adopt a simple geometric human body representation. The proposed body representation \emph{EllipBody} is composed of several ellipsoids to represent different body parts. Parameters of EllipBody include the length, thickness, and orientation of each body part. We share the pose parameters between EllipBody and SMPL models and use an Iterative Closest Points (ICP) loss and other model constraints to register SMPL on EllipBody.

Our approach contains three stages to reconstruct a human body model from coarse to fine. At the first stage, we propose an architecture to regress the parameters of EllipBody from images where supervision, including 2D keypoints and part segmentation, are predicted from pretrained networks. Although a deep neural network can take all the pixels as input, this type of one-shot prediction requires a large amount of data to form the definitive image-model mapping. To refine the predicted model, we then optimize the model by minimizing the discrepancy between 2D and projected 3D poses. Both regression and optimization processes use PartDR to compute the gradients of the vertices that derived from the residual of rendered and predicted part segmentation. Since the predicted model is well initialized by networks, optimization can be faster and easier to converge. Finally, as an additional stage, we convert the EllipBody to SMPL, transform our representation as an inscribed model of SMPL, which means we do not need to retrain the network with various templates of the detailed model. With the EllipBody and PartDR, we achieve the balanced performance for pose and shape reconstruction on Human 3.6M \cite{ionescu2014human3}, UP-3D \cite{lassner2017unite} and LSP \cite{Johnson10} datasets.

In summary, our contributions are three-fold. 
\begin{itemize}[noitemsep,topsep=0pt]
    \item We propose an occlusion-aware part-level differentiable renderer (PartDR) to make use of part segmentation as supervision for training and optimization. 
    \item We propose an intermediate human body representation (EllipBody) for human pose and shape reconstruction. It is a light-weight and part-based representation for regression and optimization tasks and can be converted to detailed models. 
    \item Our approach that contains a deep neural network together with an iterative post-optimization achieves the balanced and predominant performance in human pose and shape reconstruction.  
\end{itemize}

\section{Related Work}

\subsection{Differentiable Rendering}
Rendering connects the image plane with the 3D space. Recent works on inverse graphics \cite{de2008model,loper2014opendr,kato2018renderer} made great efforts in the differentiable property, which makes the renderer system as an optional module in machine learning. Loper \etal \cite{loper2014opendr} propose a differentiable renderer called OpenDR, which obtains derivatives for the model parameters. Kato \etal \cite{kato2018renderer} present a neural renderer which approximates gradient as a linear function during rasterization. These methods support recent approaches \cite{omran2018neural, pavlakos2018learning} exploiting the segmentation as the supervised labels to improve their performance. These differentiable renderers output the shape and textures successfully but ignore different parts of the object. Liu \etal \cite{liu2019soft} propose Soft Rasterizer, which exploits softmax function to approximate the rasterization process during rendering. However, it has several parameters to control the rendering quality that increase the number of super parameters when training. However, it has several parameters to control the rendering quality that increase the super parameters when training. We extend the differentiable renderer to part-level and propose an occlusion-aware loss function for part-level rendering. Therefore, we can explore the spatial relations between various parts in a single model that reduce the ambiguities.

\vspace{-3mm}
\subsection{Representations for Human Bodies}
The success of monocular human pose and shape reconstruction cannot stand without the parametric representations that contain the body priors. Among these representations, the 3D skeleton is a simple and effective form to represent human pose and helps numerous previous works \cite{varol2018bodynet,pavlakos2017coarse,sun2017compositional,zhou2017towards,wang2018drpose3d,xiao2018simple,habibie2019wild, xiang2019monocular}. In skeleton-based methods, joint positions \cite{martinez_2017_3dbaseline,straka2011skeletal} and volumetric heatmaps \cite{sun2018integral,pavlakos2018ordinal} are often used to predict the 3D joints. However, these methods only focus on the pose estimation while the human shape is usually disregarded. 
In order to recover the body shape together with the pose, most approaches employ parametric models to represent the human pose and shape simultaneously. These models are often generated from human body scan data and encode both pose and shape parameters. Loper \cite{loper2015smpl} propose the skinned multi-person linear model (SMPL) to generate a natural human body with thousands of triangular faces. Recently, Pavlakos \etal \cite{SMPL-X:2019} propose a detailed parametric model SMPL-X that can model human body, face, and hands altogether. These models contain thousands of vertices and faces and can represent a more detailed human body. As these parametric models use implicit parameters to indicate the human shape, it is hard to adjust body parts independently.
Many of the part-based models are derived from the work on 2D pictorial structures (PS) \cite{felzenszwalb2005pictorial,zuffi2015stitched}. 3D pictorial structures are introduced in 3D pose estimation but do not attempt to represent detailed body shapes\cite{sigal2012loose, amin2013multi, burenius20133d, belagiannis20143d}. Zuffi and Black\cite{zuffi2015stitched} propose a part based detailed model Stitched Puppet which defines a “stitching cost” for pulling the limbs apart. Our method becomes an intermediate representation for current models, representing the pose and shape of body structure with explicit parameters, and is able to be converted to any detailed model.

\vspace{-3mm}
\subsection{Human Pose and Shape Reconstruction}
Recent years have seen significant progress on the 3D human pose estimation by using deep neural networks and large scale MoCap datasets. Due to perspective distortion, body shape and camera settings also affect the pose estimation. The problem to reconstruct both human pose and shape is developed with two paradigms - optimization and regression. 
Optimization-based solutions provide more accurate results but take a long time towards the optimum. Guan \etal \cite{guan2009estimating} optimize the parameters of SCAPE \cite{anguelov2005scape} model with the 2D keypoints annotation. With reliable 2D keypoints detection, Bogo \etal \cite{bogo2016keep} propose SMPLify to optimize the parameters of SMPL model\cite{loper2015smpl}. Lassner\etal \cite{lassner2017unite} take the silhouettes and dense 2D keypoints as additional features, and extend SMPLify method to obtain more accurate results. Recent expressive human model SMPL-X \cite{SMPL-X:2019} integrates face, hand, and full-body. Pavlakos \etal optimize the VPoser \cite{SMPL-X:2019}, which is the latent space of the SMPL parameters, together with a collision penalty and a gender classifier.
The regression-based method becomes the majority trend recently because of agility. Thus Pavlakos \etal \cite{Pavlakos_2018_CVPR} use a CNN to estimate the parameters from the silhouettes and 2D joint heatmaps. Kanazawa \etal \cite{Kanazawa_2018_CVPR} present an end-to-end network, called HMR, to predict the parameters of the shape which employ a large dataset to train a discriminator to guarantee the available parameters. Kolotouros \etal \cite{kolotouros2019convolutional} propose a framework called GraphCMR which regresses the position of each vertex through a graph CNN. SPIN \cite{kolotouros2019learning} combines both optimization and regression methods to achieve high performance. Although these methods state the success of both pose and shape, and some exploit silhouettes or part segmentation as the input of neural networks, most of them only use 2D keypoints as supervision. As far as we know, our approach is the first method to utilize part segmentation as supervision, boosting the performance on both optimization-based and regression-based methods.

\vspace{-2mm}
\section{Method}\label{sec:method}

In this section, we present an algorithm to reconstruct the pose and shape of the human body from a single image. First we formulate the learning objective function (section \ref{sec:objective}). In section \ref{sec:dr}, we provide the implementation of our part-level differentiable renderer. In section \ref{sec:ellipbody}, we introduce the design of our part-level intermediate model for human bodies. The pipeline is presented in section \ref{sec:ellipnet}. As illustrated in Figure~\ref{fig:method}, the part-level differentiable renderer is integrated into an end-to-end network and an optimization loop to make use of part segmentation. 

\input{fig_method.tex}
\input{fig_partopt}
 
\vspace{-3mm}
\subsection{Learning Objective}\label{sec:objective}
Suppose a human model is controlled by pose parameters $\vec{\theta}$ and shape parameters $\vec{\beta}$. Given a generative function
\begin{equation}
\mathcal{H}(\vec{\theta}, \vec{\beta}) = (\mathcal{M}, \vec{S}).
\end{equation}
$\mathcal{M}$ is the mesh of the body, including the positions of vertices $\{\vec{v}^{i,j} \mid i=1, ..., K\}$ and the topological structure of faces $\{\vec{f}^{i,j}\mid i=1, ..., K\}$. $i$ indicates the index of body part, and $j$ indicates the index of vertex or face in the corresponding part. $\vec{S}$ is the skeleton of the mesh, consisting of the 3D positions of joints $\{\vec{s}_{i}\mid i=1, ...,N\}$. $K$ is the number of body parts, and $N$ is the number of joints.

Given a $3 \times 4$ projection matrix $\mathcal{P}$, rendering function is
\begin{equation}
    \mathcal{R}(\mathcal{M}, \vec{S}, \mathcal{P}) = (\mathcal{A}^{1},...,\mathcal{A}^{K},\vec{S}_{2D}), \label{eq:renderfunc}
\end{equation}
where $\mathcal{A}^{k}\in\mathbb{R}^{w\times h}$ is the part segmentation map of the $k$-th part. The size of rendered images are $w \times h$. $\vec{S}_{2D}$ are projected joints.

The loss function is composed of three terms, including reconstruction loss, projection loss and part segmentation loss.
\begin{equation}\label{loss_eq}
\mathcal{L}=\lambda_{3D}\mathcal{L}_{3D}+\lambda_{proj}\mathcal{L}_{proj}+\lambda_{seg}\mathcal{L}_{seg}.
\end{equation}
\vspace{-2mm}
\begin{equation}
    \mathcal{L}_{3D}=\lVert\vec{S}-\hat{\vec{S}}\rVert_{2}^{2}.
\end{equation}
\vspace{-2mm}
\begin{equation}
    \mathcal{L}_{proj}=\lVert\vec{S}_{2D}-\hat{\vec{S}}_{2D}\rVert_{1}.
\end{equation}
\vspace{-2mm}
\begin{equation}
    \mathcal{L}_{seg}=\sum_{k=1}^{K}\sum_{i=1}^{w}\sum_{j=1}^{h} \lVert\mathcal{A}_{(i,j)}^{k}-\hat{\mathcal{A}}_{(i,j)}^{k}\rVert_{2}^{2}.
\end{equation}

$\lambda_{3D}$,$\lambda_{proj}$ and $\lambda_{seg}$ are weights for each loss. We set $\lambda_{3D}=0$ for images that only have 2D annotations $\hat{\vec{S}}_{2D}$ and $\hat{\mathcal{A}}_{(i,j)}^{k}$ where $(i, j)$ is the pixel index in part segmentation.

To achieve refined pose and shape parameters, we employ a two stage framework as shown in Figure \ref{fig:method}. First we use a deep neural network $\Gamma$ to predict features.
\begin{equation}
    \Gamma(Image, \vec{W}) = (\theta, \beta, \vec{S}_{2D}, \{\mathcal{A}^{k}\}_{k=1}^{K}).
\end{equation}
$\vec{W}$ are the weights of $\Gamma$ trained by minimizing the loss in (\ref{loss_eq})
At the second stage, we optimize $\theta$ and $\beta$ by minimizing the same loss.

\subsection{PartDR: Part-Level Differentiable Renderer}\label{sec:dr}

Human part segmentation provides effective 3D evidence, e.g. boundaries, occlusions and locations, to infer the relationship between body parts. We extend previous object-level differentiable neural mesh renderer (NMR)\cite{omran2018neural} to a Part-level Differentiable Renderer (PartDR). It is applicable to deal with the multiple parts, leading each part located and posed correctly.
PartDR illustrates human parts independently and ignores the region which is occluded by other body parts during the computation of derivatives. We also design a depth-aware occlusion loss to revise the incorrectly occluded region.
As illustrated in Figure \ref{fig:partopt}, 
the target position of the red part is behind the blue one. The optimization process supervised by full segmentation fails to converge to the global optimum with a poor initialization. If supervision becomes part segmentation of two triangles, PartDR further changes the depth of them to reach the global optimum.

\input{fig_partdr.tex}

\subsubsection{Rendering the human parts}
Forward propagation of PartDR outputs the face index map $\mathcal{F}$ and the alpha map $\mathcal{A}$ in common with the traditional rendering pipeline. The face index map indicates the correspondence between image pixels and faces of the human mesh.
\begin{equation}
 \mathcal{F}(u,v)=
 \begin{cases}
\argmin_{i} \mathbf{D}_{(u,v)}(f^i_j) & \text{at least one} f^i_j; \\
-1 & \text{otherwise.}
\end{cases}
\end{equation}
where $f^i_j$ is the $j$th face of the $i$th part, and its projection covers the pixel at $(u,v)$. The function $\mathbf{D}_{(u,v)}(\cdot)$ means the depth of specified face at $(u,v)$.
The alpha map $\mathcal{A}$ is defined as a binary map that $\mathcal{A}(u,v)=1$ indicates $\mathcal{F}(u,v)>-1$. For each part, $\mathcal{A}^i$ in (\ref{eq:renderfunc}) can be calculated by $\mathcal{F}$ and $\mathcal{A}$.
\begin{equation}
\mathcal{A}^i(u,v)=
 \begin{cases}
\mathcal{A}(u,v) & \text{if} \mathcal{F}(u,v)=i; \\
0 & \text{otherwise.}
\end{cases}
\end{equation}

\subsubsection{Approximate derivatives for part rendering}

We take the insight that the differentiable rasterization is a linear function to compute approximate derivatives of each part. 
We use the function $I_{P}(x)\in{0,1}$ to indicate the rendering function of pixel $P$ at $(u, v)$, thus $I_{P} = \mathcal{A}^{i}(u,v)$.

We define $\frac{\partial I_{P}}{\partial x}(\vec{v}^{i,j}_{k})$ as the derivative of the $k$th vertex of face $f^{i}_{j}$. If the face $f^{i}_{j}$ will not be occluded by other parts,  
\begin{equation}
    \frac{\partial I_{P}}{\partial x}(\vec{v}^{i,j}_{k\in{1,2,3}})=\frac{\delta{I_{P}}}{x_0-x_1}.
\end{equation}
We only show the derivatives on the x-axis for simplicity. ${I_{P}(x)}$ is the rendered value of pixel $P$. $\delta{I_{P}}$ is the residual between ground truth $I_{P}$ and $I_{P}(x_{0})$. $x_0$ is the x-coordinate of the current vertex and $x_1$ is a new x-coordinate that makes $P$ collide the edge of the rendered face. For example, in in Figure \ref{fig:dr}(b), moving $\vec{v}_{k}^{i,j}$ from $x_0$ to $x_1$ makes $I_{P}$ from 0 to 1. We approximate this moving process as a linear transformation, so that the gradient at $x_0$ is the slope of the linear function.

Considering the circumstance that one face is occluded by others partially or completely, change in $x_1$ will not change $I_{P}$. As shown on the left side in Figure \ref{fig:dr} (a), the red triangle does not belong to the $i$-th part and occludes the blue face. The yellow pixel is $P$. Without occlusion, moving $\vec{v}_{k}^{i,j}$ from $x_0$ to $x_1$ will change the value of $P$, as illustrated by black line. If $P$ is occluded, $I_{P}$ will not change in $\mathcal{A}^{i}$, so the gradient should be 0.

\begin{equation} 
\frac{\partial I_{P}}{\partial x}(\vec{v}^{i,j}_{k\in{1,2,3}})=
\begin{cases}
0,& \text{if }\mathcal{A}(u,v)>0 \\
 & \text{and } F(u,v)\neq i;\\
\frac{\delta{I_{P}}}{x_0-x_1},& \text{otherwise.}
\end{cases}
\end{equation}

We propose derivatives on the $z$-axis (direction on depth) as an extension for the part-level neural renderer. We omit the derivatives in the occluded regions and then design a new approximation of the derivatives on the $z$-axis to refine the incorrectly occluded part. As shown in Figure \ref{fig:dr}(e), we first find the occluded face. Then we compute the depth derivatives directly proportional to the distance between the occluded point and the one occluding it. Base on the triangle similarity, the derivative is 
\begin{equation}
      \frac{\partial I_{P}}{\partial z}(\vec{v}^{i,j}_{k})=\lambda\cdot{\delta{I_{P}}}\cdot\log\left(\frac{\Delta(M_{0},Q)}{\Delta(M_{0},\vec{v}^{i,j}_{k})\cdot{\Delta{z}}}+1\right).
\end{equation}
$\Delta{z}=z_{0}-z_{1}$ is the distance between the two faces. $\Delta(\cdot, \cdot)$ is the length between two points. $Q$ is the corresponding point whose projecting point is $P$. The line form $\vec{v}^{i,j}_{0}$ to $Q$ intersects $\vec{v}^{i,j}_{1}$ to $\vec{v}^{i,j}_{2}$ at $M_{0}.$  $\lambda$ is a variable to magnitude the term. Different from $x$-axis that $x_1-x_0$ at least one pixel, $\Delta z$ could be a small value, so we use logarithmic function to avoid gradient explosion.

\subsection{EllipBody: An Intermediate Representation}\label{sec:ellipbody}

\input{fig_ellipbody.tex}

We proposed a light-weight and flexible intermediate representation to simplify the model and disentangle human body parts. The proposed representation \emph{EllipBody} utilizes ellipsoids to represent body parts, therefore pose and shape parameters can be the position, orientation, length, and thickness of each ellipsoid. The EllipBody contains both the human mesh $\mathcal{M}$ and the skeleton $\vec{S}$. 
This non-detailed model can be considered as the inscribed model of real human body and has following advantages. Firstly, the model composed of primitive geometries has fewer faces and simple typologies, which can accelerate the rendering and computation of interpenetration. Secondly, EllipBody is robust to imperfect part segmentation. Detailed models, such as SMPL, use shape parameters altogether to adjust the body. Thus all parameters may be affected by the error of one part. Lastly, in some circumstances, users may need to modify detailed models intuitively. EllipBody can be edited through its interpretable shape parameters, then SMPL will be correspondingly changed. 


As shown in Figure~\ref{fig:skeleton}, we deform an icosahedron to generate each ellipsoid then assemble them according to the bones of the skeleton, where the endpoints of the ellipsoids are located at human joints. We define the shape of the $i$-th ellipsoid $\mathcal{E}_i$ with the bone length $l_i$ and part thickness along two other axes $t^1_i, t^2_i$. The global rotation $\vec{R_i}\in \mathbf{SO}(3)$ of each ellipsoid is the pose parameter. In practice, EllipBody is made of 20 ellipsoids:
\begin{equation}
    \mathbb{E}=\{\mathcal{E}_{i}|i=1,...,20\},
\end{equation}
where
\begin{equation}
\mathcal{E}_{i}=E(\vec{R_i}, \vec{C_i}, l_i, t_i^1, t_i^2) \label{ellipsoid_eq}.
\end{equation}
$E$ is the function to deform the original icosahedron. $\vec{C_i}$ is the position of the centre of the $i$-th ellipsoid, which is the midpoint of two connected joints. By forward kinematics, $\vec{S}$ is inferred via $l$.
\begin{equation}
\vec{S}_{i}=\vec{R}_{parent(i)}\cdot (l_{i}\vec{O}_{i})  + \vec{S}_{parent(i)}.
\label{skeleton_eq}
\end{equation}
where $\vec{O}_{i}$ is an offset vector indicating the direction from its parent to the current joint. So $l_{i}\vec{O}_{i}$ denotes the local position of $i$-th joint in its parent joint's coordinate. 

As the human body is symmetric, ellipsoids in EllipBody share the parameters when indicating the same category of human parts. Therefore, we reduce the number of semi-principal axes parameters from $\mathbb{R}^{20 \times 3}$ to $\mathbb{R}^{27}$. Table \ref{tab:params} shows the indices of shape parameters for different parts.

\input{supp_param_ellip.tex}

EllipBody can be an intermediate representation for any detailed models. We use SMPL\cite{loper2015smpl} as an example to describe the conversion between them. We consider that EllipBody should inscribe the SMPL model when they represent the same objective. Suppose $\vec{v}^{S}$ are vertices of the SMPL $\mathcal{M}^{S}$. $\vec{v}^{E}$ and $\mathcal{M}^{E}$ are for EllipBody. Because two models share the same pose parameters and different shape parameters, we minimize the following loss:
\begin{equation}\label{eq:ellip2smpl}
    (\theta, \beta) = \argmin_{\theta, \beta} \mathcal{L}_{ICP}(\mathcal{M}^{S}, \mathcal{M}^{E}) + \mathcal{L}_{PEN}(\vec{v}^S, \mathcal{M}^{E}), 
\end{equation}
where $\theta$ and $\beta$ are the SMPL parameters. $\mathcal{L}_{ICP}$ is Iterative Closest Points (ICP)\cite{besl1992method} loss to make two meshes closer. $\mathcal{L}_{PEN}$ is the penetration loss indicating the extent that vertices of SMPL penetrate into EllipBody. 
Suppose one vertex of SMPL $\vec{v} = (\vec{v}_{x}, \vec{v}_{y}, \vec{v}_{z})^{T}$ and one ellipsoid of EllipBody $\mathcal{E}_{i}$. The vector from the center of the ellipsoid to the vertex before rotation is $\vec{d}=\vec{R}_{i}^{-1}(\vec{v}-\vec{C}_{i})$. We transform the ellipsoid to a sphere with radius 1, and 
\begin{equation}
    e(\vec{v}, \mathcal{E}_{i}) = \left \| \frac{2\cdot\vec{d}_{x}}{l_{i}}, \frac{2\cdot\vec{d}_{y}}{t_{i}^{1}}, \frac{2\cdot\vec{d}_{z}}{t_{i}^{2}}\right \|_{2}
\end{equation} is the distance from the center to the vertex.

\begin{equation}
    \mathcal{L}_{PEN} = \sum_{\vec{v}, \mathcal{E}}{a(1- e(\vec{v}, \mathcal{E}_{i}))^{2}}. 
\end{equation}
If $e(\vec{v}, \mathcal{E}_{i}) > 1$, which means $\vec{v}$ is outside the $\mathcal{E}_{i}$, $a$ becomes 0 or otherwise 1.

\subsection{Pose and Shape Reconstruction}\label{sec:ellipnet}
We proposed an end-to-end pipeline to estimate the parameters of EllipBody. As shown in Figure \ref{fig:method}, a CNN-based backbone extracts the features from a single image first. Base on the image feature, we regress the parameters of the pose and shape. After that, we optimize the model by minimizing the loss in (\ref{loss_eq}) to achieve more accurate results. At last, we convert EllipBody to a detailed model to present the final result.
As the previous works have great success in training a Deep CNN for human pose estimation, we take ResNet-50 \cite{he2016deep} as our encoder to extract the features from an image. There are two head networks to predict 2D keypoints\cite{xiao2018simple} and body part segmentation \cite{omran2018neural}.
We use the multilayer perceptron (MLP) to regress EllipBody parameters, which consists of linear layers, batch normalization layers, ReLU, and dropouts, iterated with the residual connection. It outputs ${\vec{r}\in\mathbb{R}^{20 \times 3}}$, $\vec{l}\in\mathbb{R}^{12}$, $\vec{t}\in\mathbb{R}^{15}$, and $\vec{c}\in\mathbb{R}^{3}$. Here $\vec{r}$ is local rotation vectors, thus we can compute the global rotation $\vec{R}\in\mathbb{R}^{20 \times 3}$ by forward kinematic \cite{lee1988kinematic,liu1993kinematic} of EllipBody. Note that output $\vec{r}$ may not be available rotations, so we employ Gram–Schmidt process \cite{daniel1976reorthogonalization} to ensure the plausibility. Camera parameters $\vec{c}$ for a weak-perspective model are proposed by Kanazawa \etal \cite{Kanazawa_2018_CVPR}.
The skeleton $\vec{S}$ and the vertices of mesh can be calculated from $\vec{r}$, $\vec{l}$, and $\vec{t}$ as mentioned in Section \ref{sec:ellipbody}. Given the camera parameters and EllipBody, PartDR outputs $K$ part maps $\{\mathcal{A}^1, ..., \mathcal{A}^K \}$ and projected joints $\vec{S_{2D}}$.

The neural network could map all the images and human models in theory. However, due to lack of 3D annotations, the performance of the network may not be satisfied. To achieve more accurate results, we take the predicted parameters from the regressor as the initialization, then use part segmentation from the network to refine the EllipBody. We minimize the loss function
\begin{equation}
    E(\vec{r}, \vec{l}, \vec{t}, \vec{c}) = \lambda_{seg}\mathcal{L}_{seg} + \lambda_{proj}\mathcal{L}_{proj} + \lambda_{l}\mathcal{L}_{l} + \lambda_{t}\mathcal{L}_{t}.
\end{equation}
 Most terms are introduced in section \ref{sec:objective}. $\mathcal{L}_{l}=\lVert l - \Bar{l} \rVert^{2}$ and $\mathcal{L}_{t}=\lVert t - \Bar{t} \rVert^{2}$ are regularization term where $\Bar{l}$ and $\Bar{t}$ are mean shape parameters.
Since the network provides an initialization, the optimization process only needs a small number of iterations to converge. As a result, the optimization fixes the minor errors of the model predicted from regressors.

In order to visualize the detailed body model, we fit SMPL model to circumscribe the EllipBody. Note that SMPL has 23 rotation angles, while ours have 20 angles. We decompose the angles where the angle in SMPL are almost fixed like the chest and clavicles. Other joints related to limbs and the head share the same pose parameters. After that, we initialize $\beta$ by mean shape parameters and minimize the loss function proposed in (\ref{eq:ellip2smpl}).

\section{Experiments}
In this section, we show the evaluation of the results achieved by our method and perform experiments with ablation studies to analyze the effectiveness of the components in our framework. We also compare our method with state-of-the-art pose and shape reconstruction techniques. The qualitative evaluation adopts SMPL\cite{loper2015smpl} as the final model to keep the consistency with other methods, despite that any other model can be converted from EllipBody with the same scheme.

\subsection{Datasets}
   
\textbf{Human3.6M:} It is a large-scale human pose dataset that contains complete motion capture data which contain rotations angles. It also provides images, camera settings, part segmentation, and depth images. We use original motion capture pose data to train EllipBody and assemble the body part segmentation into 14 parts. We use data of five subjects (S1, S5, S6, S7, S8) for training and the rest two (S9, S11) for testing. We use the Mean Per Joint Position Error (MPJPE) and Reconstruction Error (PA-MPJPE) for evaluation.

\noindent\textbf{UP-3D:} It is an in-the-wild dataset, which collects high-quality samples from MPII \cite{andriluka20142d}, LSP \cite{Johnson10}, and FashionPose \cite{dantone2014body}. There are 8,515 images, 7,818 for training and 1389 for testing. Each image has corresponding SMPL parameters. We use this dataset to enhance the generalization of the network.

\noindent\textbf{LSP:} It is a 2D pose dataset, which provides part segmentation annotations. We use the test set of this dataset to evaluate the accuracy and f1 score of part segmentation. 

\subsection{Implementation Details}
The backbone ResNet-50 and 2D keypoints estimation network are pretrained by Xiao \etal \cite{xiao2018simple}. The dimension of the regression model is 1024, and each regressor is stacked with two residual blocks. The input image size is $256\times 256$, while output size of segmentation is also $256\times256$. We use Adam optimizer for network training and optimization. 
During the training process, we use Human3.6M data to warm up. It is an important fact that the somatotype of body and extrinsic camera parameters are coupling due to perspectives. Therefore, we adopt mean bone lengths $\Bar{\vec{l}}$ and thicknesses $\Bar{\vec{t}}$, and train the regressors for $\vec{r}$, $\vec{c}$. we set the batch size to 128, with learning rate $10^{-3}$ to train the model for first 70 epochs without part segmentation loss $\mathcal{L}_{seg}$. At the second stage, we add segmentation loss and reduce the learning rate to $10^{-4}$ for additional 30 epochs. We use UP-3D and Human3.6M together and fix the weights of regressor for $\vec{c}$.
During optimization, we use Adam optimizer with learning rate $10^{-2}$ to fit the predicted EllipBody at most 50 iterations. Base on the experimental results, we set $\lambda_{3D}:\lambda_{proj}:\lambda_{seg}:\lambda_{l}:\lambda_{t}=1:1:10^{-2}:10^{-3}:10^{-3}$. 

\subsection{Comparing with the State-of-the-Arts.}\label{sec:sota}

\input{exp_h36m_sota.tex}
\input{exp_up3d_shape}

We compare our approach with other state-of-the-art methods for both pose and shape reconstruction. For 3D pose estimation, we evaluate the results with Human3.6M dataset in Table~\ref{tab:hm36m}. We use the same protocols in \cite{Kanazawa_2018_CVPR} where the metric is reconstruction error after scaling and rigid transformation. In order to evaluate the results for body shape reconstruction, we compare our method with recent approaches on both 2D and 3D metrics. The results in Table~\ref{tab:up3d} present per vertex reconstruction error on UP-3D. We also evaluate the results on the test set of LSP in Table~\ref{tab:lsp} with the metric accuracy and F1 score. We adopt EllipBody as the representation to avoid the minor error caused by the detailed human template.

We refer the network prediction in our approach as EllipBodyNet. The EllipBody parameters predicted directly from EllipBodyNet can achieve competitive results among recent methods which reconstruct the full body in an end-to-end network. 
With further optimization, the error is close to the state-of-the-art approach \cite{kolotouros2019learning}, which also conducts the optimization to improve the results. We justify that our method balances the pose and shape. As shown in Table~\ref{tab:lsp}, with predicted part segmentation from \cite{omran2018neural}, optimized EllipBody outperforms others including SPIN \cite{kolotouros2019learning} on both foreground/background segmentation and part segmentation. The rationale is that those optimization targets in previous methods\cite{bogo2016keep, kolotouros2019learning} are typically 2D keypoints. Leveraging part segmentation as the optimization goal not only provides faithful shape parameters but also improves the performance of the pose estimation as well. We find that optimization with ground-truth part labels even out-perform SMPLify\cite{bogo2016keep} who generates the part annotations for the UP-3D dataset. For a comprehensive comparison, we also optimize SMPL model on ground truth part segmentation. The result is roughly equivalent to EllipBody but does not perform better on part segmentation due to the heterogeneous mesh of SMPL template. Since UP-3D dataset has ground truth SMPL parameters, we evaluate registered SMPL results with vertex-to-vertex errors in Table~\ref{tab:up3d}.



\subsection{Ablative Study}
\input{exp_lsp_iou}
\input{exp_h36m_ablative.tex}

\noindent\textbf{Model Representation.}
We investigate the effectiveness of our proposed model base on protocol 1 of Human3.6M. Different from section \ref{sec:sota}, this protocol only aligns the root of the models to compute MPJPE. The reason to adopt this protocol is that our EllipBodyNet predicts both pose and shape parameters, and scaling or applying rigid transformation on the model will neglect the global rotation and body somatotype.
We first compare the SMPL and EllipBody as the output of the network, respectively. While 3D annotations such as joint angle, shape parameters are not precisely the same between two models, we train the network supervised by 2D keypoints and full/part segmentation. As shown in Table \ref{table:ablation}, the performance is powered by EllipBody, which is composed of primitive ellipsoids that facilitate the training process.


\noindent\textbf{Body Part Supervision.}
To prove that part segmentation provides more image cue, we take various loss functions to train the model respectively. Full body segmentation reduces error by $6.7 mm$, comparing with projection only. When we replace with part supervision, the result decreases further $1.9 mm$ MPJPE. Even if we train the network with 3D supervisions, part segmentation still improves the performance of the human pose and shape reconstruction, as shown in Table \ref{table:ablation}.

\noindent\textbf{Speed of Part Differentiable Renderer.}
Body parts of EllipBody are deformed from the icosahedron, which means we can subdivide each face to provide more fine-grained EllipBody. In Figure~\ref{fig:optimize}, $E_{i}$ denotes the EllipBody which is subdivided by $i$ times. Each face is divided into four identical faces at each time. With more faces, the model can fit the body boundary more compactly, but consume more time and may lead to over-fitting. The red line illustrates the iteration time for different models, and the blue line indicates the accuracy of part segmentation on LSP\cite{Johnson10}. We find that the performance will not increase after two subdivisions, so we adopt $E_1$ as the default EllipBody configuration. The speed is tested on GTX 1080TI GPU, and the size of the rendered image is $256 \times 256$.
\input{fig_speed_smpl_ellip}

\subsection{Qualitative Evaluation}
Figure~\ref{fig:qualitative} illustrates the reconstruction results by our approach in LSP dataset. We present the part segmentation rendered by PartDR, and the mesh of both EllipBody and corresponding SMPL. The second and third columns show the results by recent the state-of-the-art methods~\cite{kolotouros2019convolutional, kolotouros2019learning}. Most pose ambiguities with precise part segmentation can be solved with PartDR and EllipBody as listed in the last column. The part segmentation is rendered with the parts of EllipBody by PartDR. The boundaries of EllipBody are close to the one in original images, proving that the non-detailed model could fit the real human body successfully.
However, there are still a few failure cases that can be attributed to incorrect part segmentation, occlusions by multiple people, as well as challenge poses. As shown in Figure~\ref{fig:failure}, though part segmentation with correct ordinal depth, the legs get crossing due to the wrong size of the part segmentation of right foot.

Another advantage of EllipBody is interpretable shape parameters. For those unusual body shapes, EllipBody can construct longer limbs, more oversized heads, as well as shorter legs. Figure~\ref{fig:smpl} shows an interpolation from one EllipBody to another. We also convert the model to SMPL. It can be widely used for animation editing and avatar modelling. 

\input{fig_failure}
\input{fig_somatotype.tex}

\input{fig_qualitative.tex}

\section{Conclusion}
In this paper, we focus on improving monocular human pose and shape reconstruction with a common image cue - part segmentation. We propose an intermediate geometry model EllipBody which has explicit parameters and light-weight structure. We also propose a differentiable mesh renderer to a depth-aware part-level renderer that can recognize the occlusion between human parts. The part-level differential neural renderer utilizes the part segmentation as effective supervision to improve the performance on both regression and optimization stage. Furthermore, any detailed parametric model like SMPL can be registered on EllipBody model, such that the neural network to predict EllipBody does not need retraining when using a new human template. Although our method still has some limitations such as uncontrollable face directions, self interpenetration and the heavy dependence of part segmentation quality, the EllipBody has great potential in many applications. As the future work, we would extend the approach to the scene involving multiple people, as well as other occlusion cases.
 
\section*{Acknowledgements}
We thank for the support from National Natural Science Foundation of China (61972157, 61902129), Shanghai Pujiang Talent Program (19PJ1403100), Economy and Information Commission of Shanghai (XX-RGZN-01-19-6348), National Key Research and Development Program of China (No. 2019YFC1521104). We also thank the anonymous reviewers for their helpful suggestions.
\bibliographystyle{eg-alpha-doi} 
\bibliography{main}       



\end{document}

%% file: fig_motivation.tex
\begin{figure}[t]
	\begin{center}
		\includegraphics[width = \linewidth]{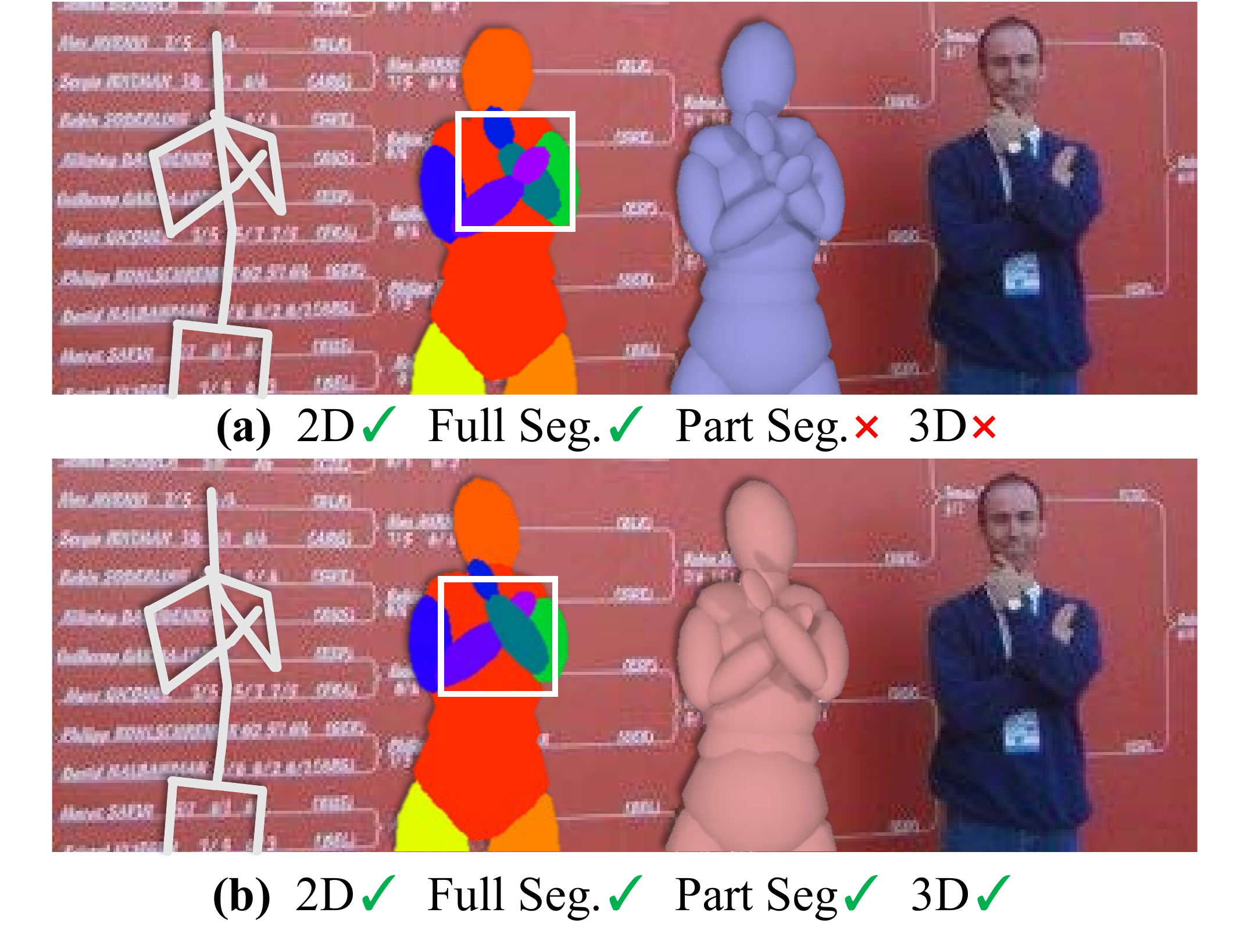}
	\end{center}
	\caption{\textbf{Effect of part segmentation.} Predictions of 2D pose and full segmentation are identical in \textbf{(a)} and \textbf{(b)}. The only difference between them is part segmentation in the white box, which indicates the different pose of crossed arms. Therefore, \textbf{(b)} with correct part segmentation performs the proper reconstruction rather than \textbf{(a)}.}
	\label{fig:motivation.}
	\vspace{-5mm}
\end{figure}

%% file: fig_method.tex
\begin{figure*}[t]
	\begin{center}
		\includegraphics[width = .95\textwidth]{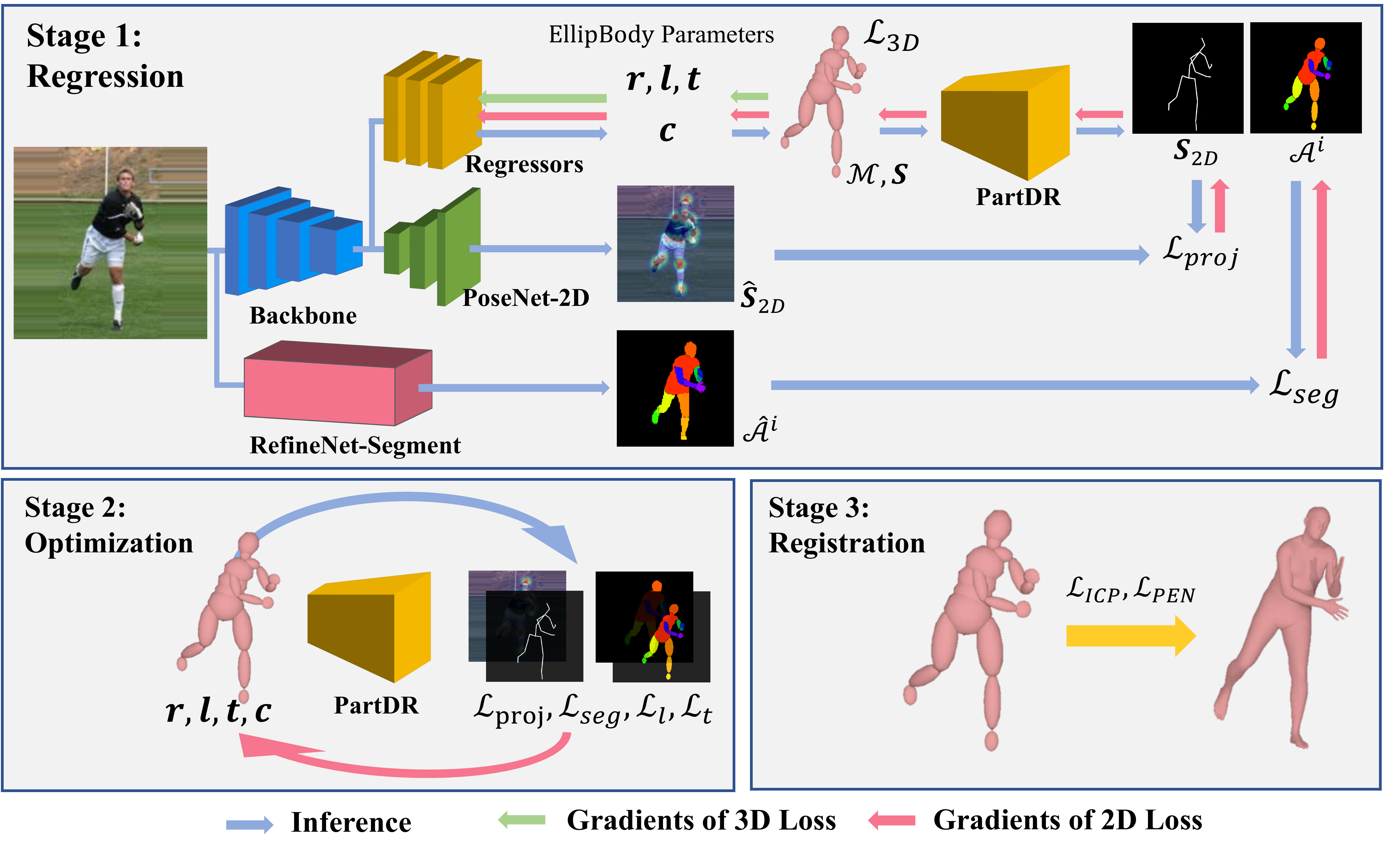}
	\end{center}
	\vspace{-4mm}
	\caption{\textbf{Framework.} Our pipeline is divided into three stages: (1) A CNN-based backbone extracts features from the image, then regressors predict the parameters of EllipBody and camera. The supervision includes 2D keypoints and part segmentation predicted from networks. (2) Using PartDR further optimizes EllipBody to align with the predicted image cues. (3) Registering a detailed model on EllipBody. }
	\vspace{-4mm}
	\label{fig:method}
\end{figure*}

%% file: fig_partopt.tex
\begin{figure}[tbp]
	\begin{center}
		\includegraphics[width = \linewidth]{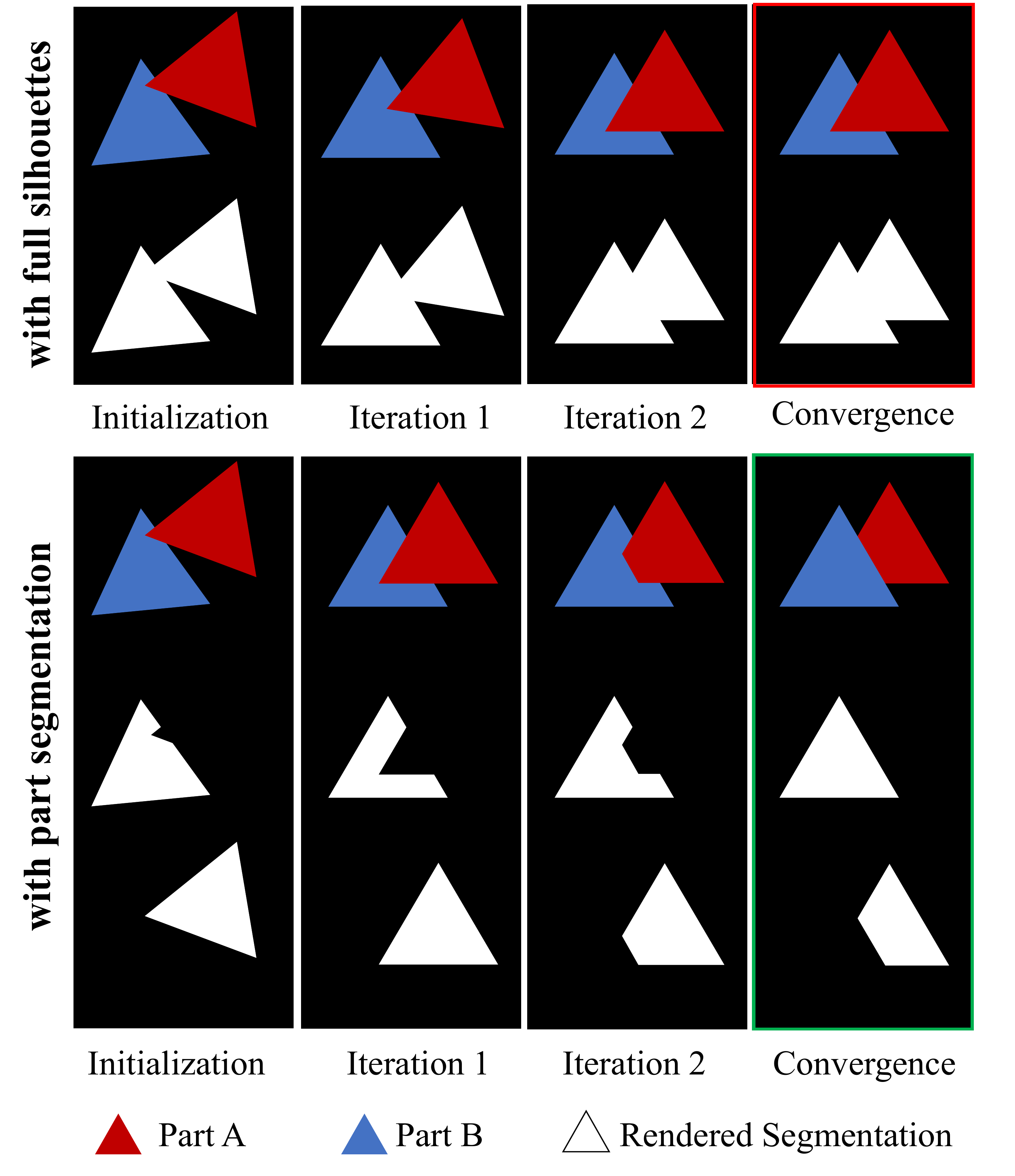}
	\end{center}
	\caption{\textbf{Optimization with PartDR.} The upper row illustrates the optimization process with full segmentation. The lower row shows the one with part segmentation. The optimization target is in the green box. However, optimization with full segmentation may stop at red box, while PartDR further pushes Part A to the right place.}
	\vspace{-6mm}
	\label{fig:partopt}
\end{figure}

%% file: fig_partdr.tex
\begin{figure*}[htbp]
    \vspace{-2mm}
	\begin{center}
		\includegraphics[width = .9\textwidth]{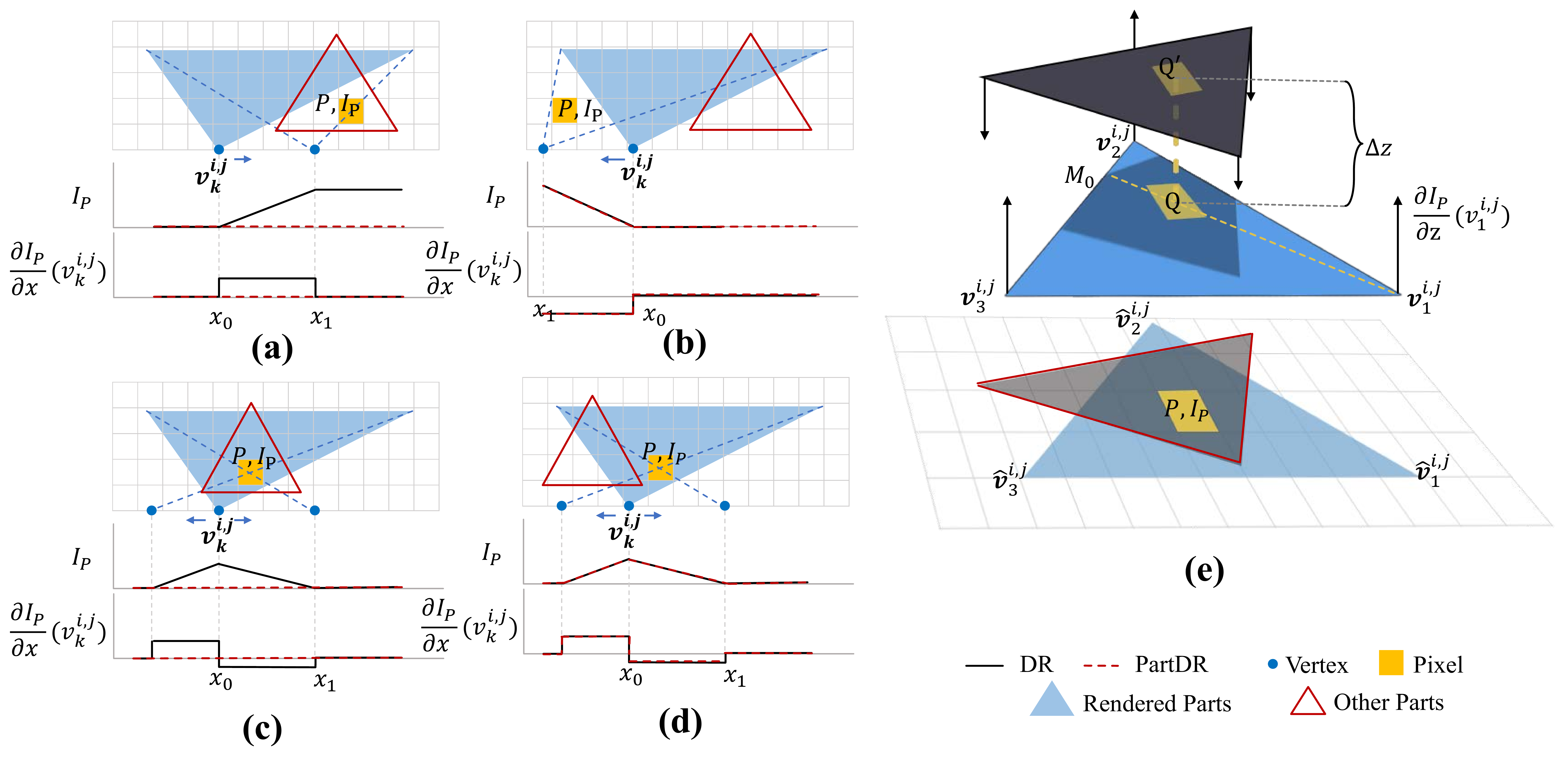}
	\end{center}
	\vspace{-3mm}
	\caption{\textbf{Part-level Differentiable Renderer.} The left side illustrates all four possible cases on approximate values of intensity and their corresponding derivatives along the axes in the image plane. The right side illustrates the case where rendered parts are occluded incorrectly, i.e.\ $z$-axis gradients take effect over the rendered parts. The yellow pixel $P$ indicates where the rendering loss come from. The black line is the approximate function of $I_P$ and its derivative if there is no occlusion. The red line is the function that occluded by red triangle. }
	\vspace{-3mm}
	\label{fig:dr}
\end{figure*}

%% file: fig_ellipbody.tex
\begin{figure} [t]
	\begin{center}
		\includegraphics[width = \linewidth]{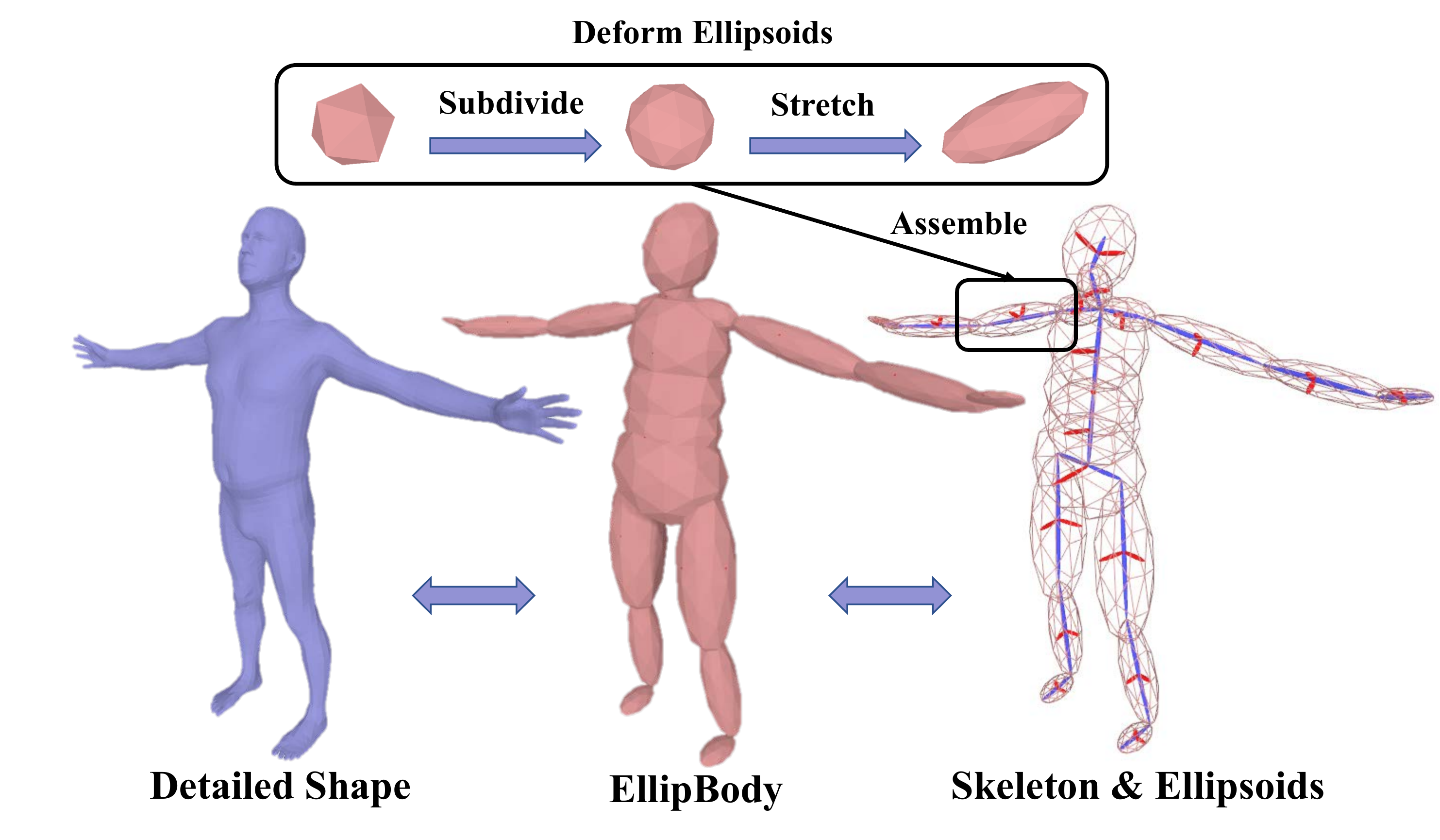}
	\end{center}
	\caption{\textbf{Formulation of EllipBody.} Ellipsoids are subdivided and deformed from icosahedrons. The skeleton of the human body is computed by forward kinematics. After that, ellipsoids are assembled along the skeleton to construct the mesh. Detailed models can be registered by optimization with ICP and penetration loss.}
	\label{fig:skeleton}
\end{figure}

%% file: supp_param_ellip.tex


\begin{table}[tb]
\vspace{-2mm}
\centering
\small
\resizebox{\linewidth}{!}{%
\begin{tabular}{c|c|cc|c|c|cc} 
\toprule

Part        & length & \multicolumn{2}{c|}{Shape} & Part & length & \multicolumn{2}{c}{Shape}     \\ 
\midrule
Ass         & $l_{0}$  & $t_{0}$ & $t_{1}$   & Upper legs  & $l_{6}$  & $t_{7}$  & $t_{7}$  \\
Abdomen     & $l_{1}$  & $t_{0}$ & $t_{2}$   & Lower legs  & $l_{7}$  & $t_{8}$  & $t_{8}$  \\
Chest       & $l_{2}$  & $t_{0}$ & $t_{3}$   & Feet        & $l_{8}$  & $t_{9}$ & $t_{10}$  \\
Neck        & $l_{3}$  & $t_{4}$ & $t_{4}$   & Upper arms  & $l_{9}$  & $t_{11}$ & $t_{11}$ \\
Shoulders   & $l_{4}$  & $t_{5}$ & $t_{5}$   & Fore arms   & $l_{10}$ & $t_{12}$ & $t_{12}$ \\
Head        & $l_{5}$  & $t_{6}$ & $t_{6}$   & Hands       & $l_{11}$ & $t_{13}$ & $t_{14}$ \\ 

\bottomrule
\end{tabular}}
\vspace{0.5mm}
\caption{\textbf{Shape Parameters of EllipBody Parts.} $\vec{l}$ denotes the length of the ellipsoids. $\vec{t}$ denotes the thickness of them. 
}
\vspace{-4mm}
\label{tab:params}
\end{table}

%% file: exp_h36m_sota.tex
\begin{table}[tbp]
\centering
\small
\hspace{-3mm}
\tabcolsep=1.75mm
\begin{tabular}{p{6cm} c}
\toprule
 & Rec. Error ($mm$)\\
\midrule
Akhter \& Black~\cite{akhter2015pose} & 181.1\\
Ramakrishna~\etal~\cite{ramakrishna2012reconstructing} & 157.3\\
Zhou~\etal~\cite{zhou2017sparse} & 106.7\\
SMPlify~\cite{bogo2016keep} & 82.3\\
Lassner~\etal~\cite{lassner2017unite} & 80.7\\
Pavlakos~\etal~\cite{Pavlakos_2018_CVPR} & 75.9 \\
NBF~\cite{omran2018neural} & 59.9 \\
HMR~\cite{Kanazawa_2018_CVPR} & 56.8 \\
GraphCMR~\cite{kolotouros2019convolutional}&50.1 \\
DenseRaC~\cite{xu2019denserac}&48.0 \\
SPIN~\cite{kolotouros2019learning}&\textbf{41.1} \\
\hline
EllipBodyNet & 49.4 \\
EllipBodyNet + Optimization & 45.2 \\
\bottomrule
\end{tabular}
\vspace{1.5mm}
\caption{Quantitative results on Human3.6M~\cite{ionescu2014human3}. Numbers are reconstruction errors (mm) of 17 joints on frontal camera (camera 3). The numbers are taken from the respective papers.}
\label{tab:hm36m}
\end{table}

%% file: exp_up3d_shape.tex
\begin{table}[tbp]
\centering
\small
\hspace{-3mm}
\tabcolsep=1.75mm
\begin{tabular}{p{3cm} c}
\toprule
 & Vertex-to-Vertex Error ($mm$)\\
\midrule
Lassner~\etal~\cite{lassner2017unite} & 169.8 \\
Pavlakos~\etal~\cite{Pavlakos_2018_CVPR} & 117.7 \\
Bodynet~\etal~\cite{varol2018bodynet} & 102.5 \\
\hline
Ours & 100.4 \\
\bottomrule
\end{tabular}
\vspace{1.5mm}
\caption{Vertex-to-Vertex error on UP-3D. The numbers are mean per vertex errors (mm). We use the registered SMPL model of Stage 3 due to the ground truth are based on SMPL models.}
\label{tab:up3d}
\end{table}

%% file: exp_lsp_iou.tex
\begin{table}[tbp]

	\centering
	\small
	\hspace{-3mm}
	\tabcolsep=2.95mm
	\begin{tabular}{@{}lcccc@{}}
		\toprule
		& \multicolumn{2}{c}{FB Seg.} & \multicolumn{2}{c}{Part Seg.} \\
		\cmidrule{2-5}
		& acc. & f1 & acc. & f1 \\
		\midrule
		SMPLify on GT~\cite{bogo2016keep} & 92.17 & 0.88 & 88.82 & 0.67 \\
		SMPLify~\cite{bogo2016keep} & 91.89 & 0.88 & 87.71 & 0.64 \\
		SMPLify on~\cite{pavlakos2018learning} & 92.17 & 0.88 & 88.24 & 0.64 \\
		HMR~\cite{Kanazawa_2018_CVPR} & 91.67 & 0.87 & 87.12 & 0.60 \\
		Bodynet \cite{varol2018bodynet} & \textbf{92.75} & 0.84 & - & -\\
		DenseRaC \cite{xu2019denserac} & 92.40& 0.88 & 87.90 & 0.64 \\
		GarphCMR~\cite{kolotouros2019convolutional} & 91.46 & 0.87 & 88.69 & 0.66 \\
		SPIN~\cite{kolotouros2019learning} & 91.83& 0.87 & 89.41 & 0.68 \\
		\midrule
		PartDR+SMPL+GT.Part & 94.03 & 0.91 & 91.91 & 0.79\\
		PartDR+EllipBody+GT.Part & 94.74 & 0.92 & 93.26 & 0.84\\
		\midrule
		PartDR+EllipBody+Pred.Part &92.13&  \textbf{0.88}& \textbf{90.70} &\textbf{0.74} \\
		\bottomrule
	\end{tabular}
	\vspace{1.5mm}
	\caption{Segmentation evaluation on the LSP. The numbers are accuracies and f1 scores of foreground-background segmentation and body part segmentation. SMPL or EllipBody indicates which body representation we used. We show the upper bound of our approach with ground truth part segmentation. Our approach with predicted part segmentation reaches the state-of-the-art. 
	}
	\label{tab:lsp}
\end{table}

%% file: exp_h36m_ablative.tex
\begin{table}[tbp]
    \vspace{0.5mm}
	\centering
	\small
	\begin{tabular}{llc}
		\toprule
		Model&Loss & MPJPE ($mm$) \\
		\midrule
		SMPL & $\mathcal{L}_{proj}$ & 106.8 \\
		SMPL & $\mathcal{L}_{proj}+\mathcal{L}_{seg}$(full) & 75.9 \\
		SMPL & $\mathcal{L}_{proj}+\mathcal{L}_{seg}$(part)&67.1 \\
		\midrule
		EllipBody & $\mathcal{L}_{proj}$ & 73.8 \\
		EllipBody & $\mathcal{L}_{proj}+\mathcal{L}_{seg}$(full)&67.1 \\
		EllipBody & $\mathcal{L}_{proj}+\mathcal{L}_{seg}$(seg)& 65.2 \\
		EllipBody & $\mathcal{L}_{3D}+\mathcal{L}_{proj}+\mathcal{L}_{seg}$(full) & 64.1 \\
		EllipBody&$\mathcal{L}_{3D}+\mathcal{L}_{proj}+\mathcal{L}_{seg}$(part)&  \textbf{62.8}\\
		\bottomrule
	\end{tabular}
	\vspace{1.5mm}
	\caption{Comparison on the protocol 1 of Human3.6M with different parametric model. We also compare the performance with or without different losses. The evaluation method is MPJPE.}
	\label{table:ablation}
	\vspace{-5mm}
\end{table}

%% file: fig_speed_smpl_ellip.tex
\begin{figure}[tbp]
	\begin{center}
		\includegraphics[width = .5\textwidth]{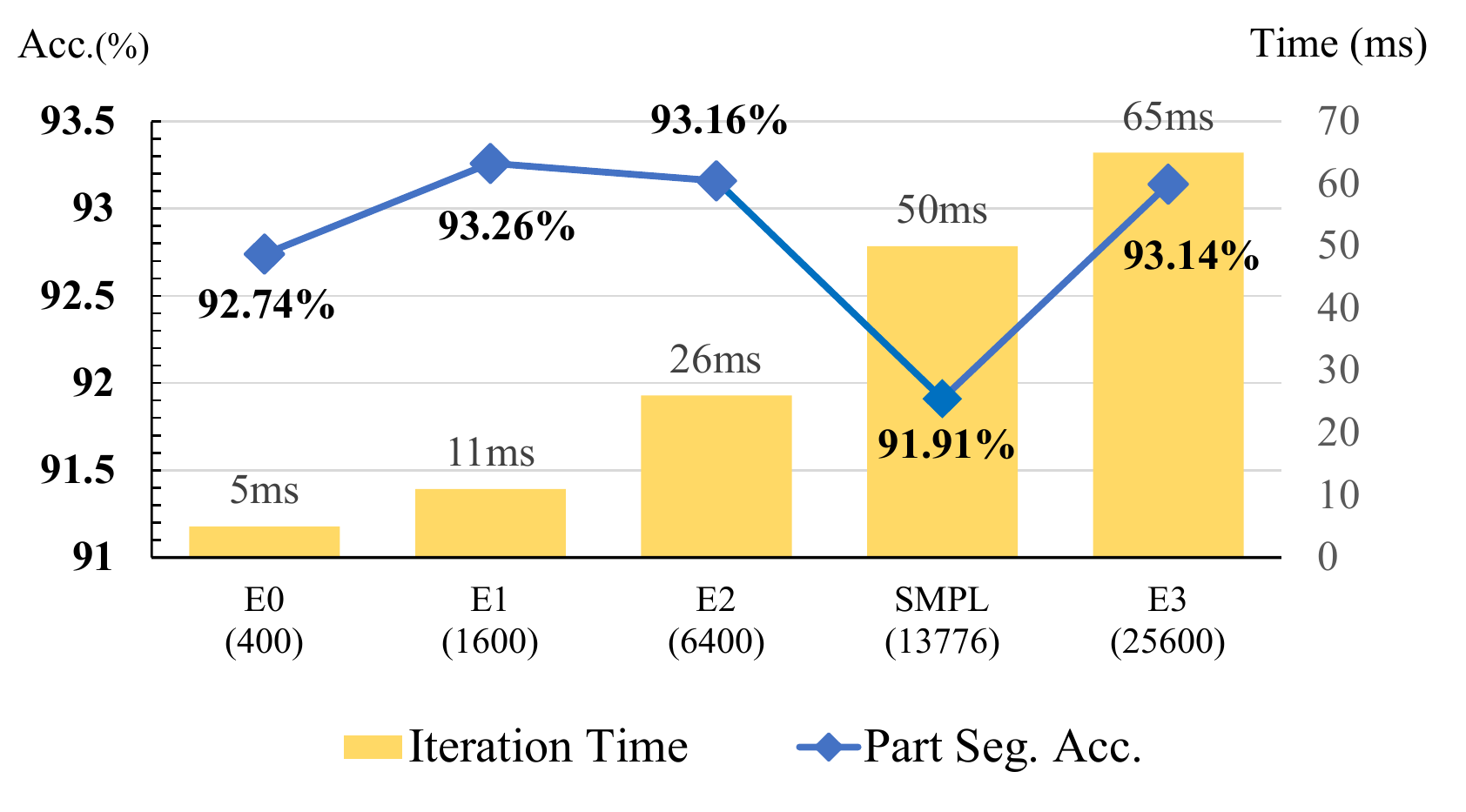}
	\end{center}
	\caption{\textbf{Optimization Performance on LSP.} The red line illustrates the time consumption of each iteration. The blue line is the accuracy of part segmentation. We mark the number of faces for our models and SMPL, and our models with repeat times of surface subdivisions from zero to three are denoted by terms from $E0$ to $E3$. The accuracy does not increase after one times subdivision.
	}
	\label{fig:optimize}
\end{figure}

%% file: fig_failure.tex
\begin{figure}[t]
	\begin{center}
		\includegraphics[width = \linewidth]{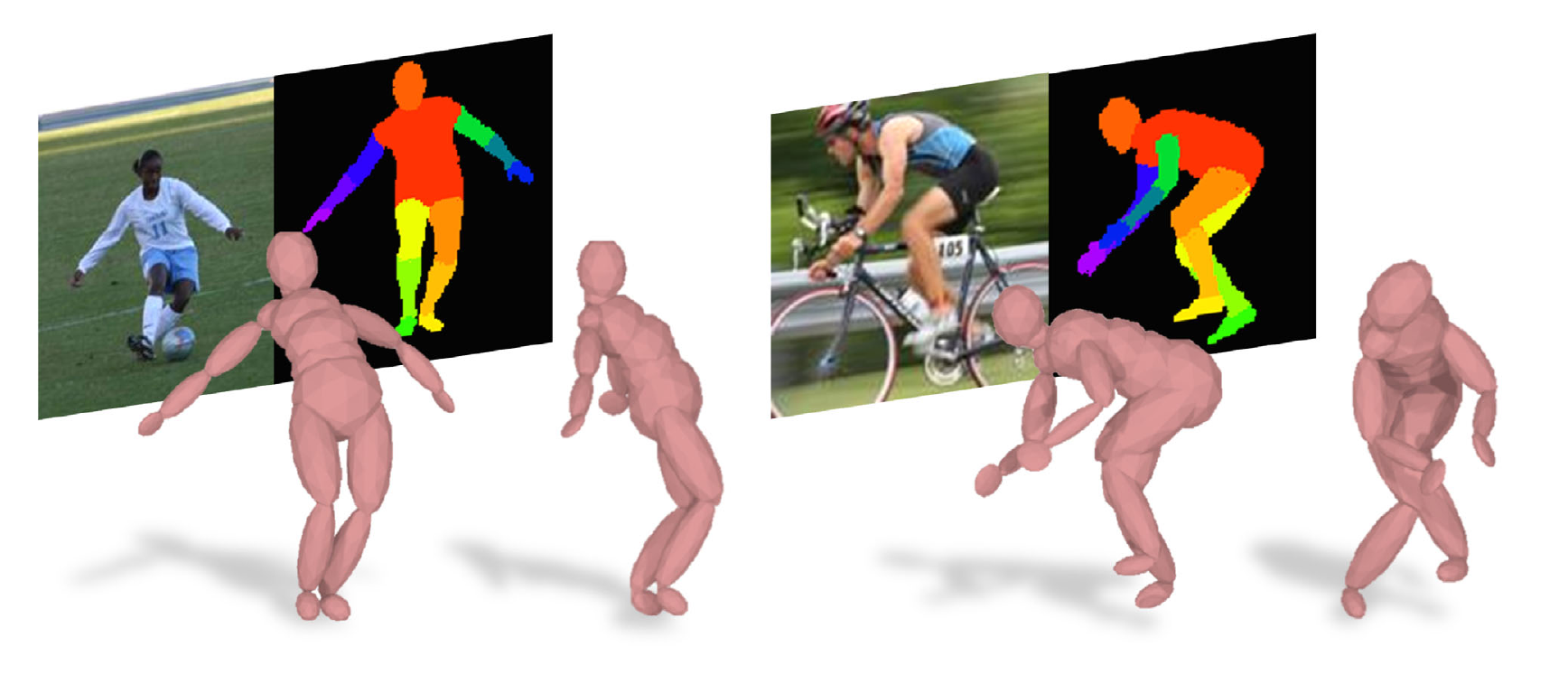}
	\end{center}
	\caption{Some failure cases of our reconstruction results are caused by wrong part segmentation and ill-posed problems. The human on the left side should be lean forward, but ours leans back. On the right side, the part segmentation of the right foot is too big, making the legs crossed. Both of them are correct at the frontal view.}
	\label{fig:failure}
\end{figure}

%% file: fig_somatotype.tex
\begin{figure}[t]
\vspace{3mm}
	\begin{center}
		\includegraphics[width = \linewidth]{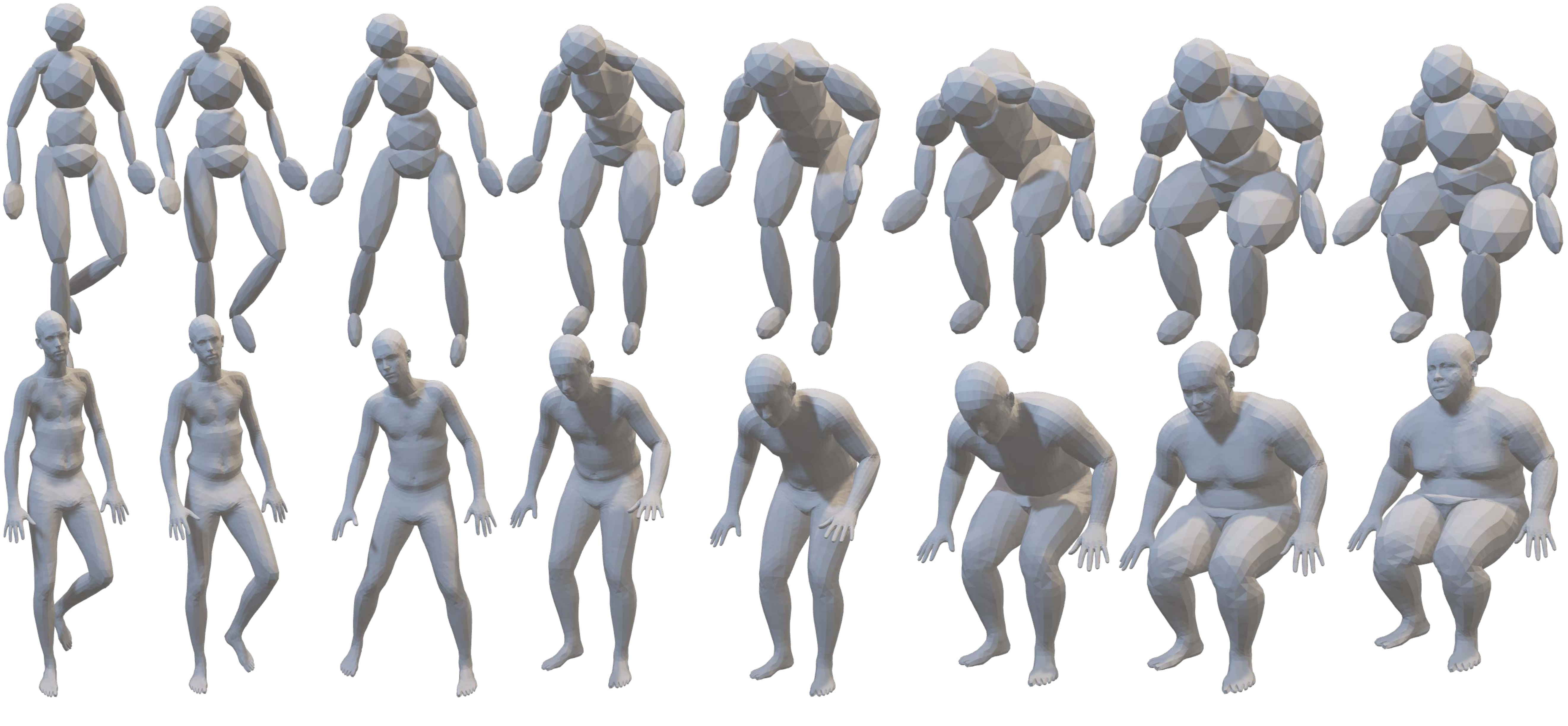}
	\end{center}
	\caption{\textbf{SMPL with EllipBody.} As the high flexibility in parameters of EllipBody, we can control the body parts independently. The upper row shows an interpolation between two totally different EllipBody models. The lower row shows the corresponding SMPL.}
	\label{fig:smpl}
\end{figure}

%% file: fig_qualitative.tex
\begin{figure*}[tbp]
	\begin{center}
		\includegraphics[height=21cm,keepaspectratio]{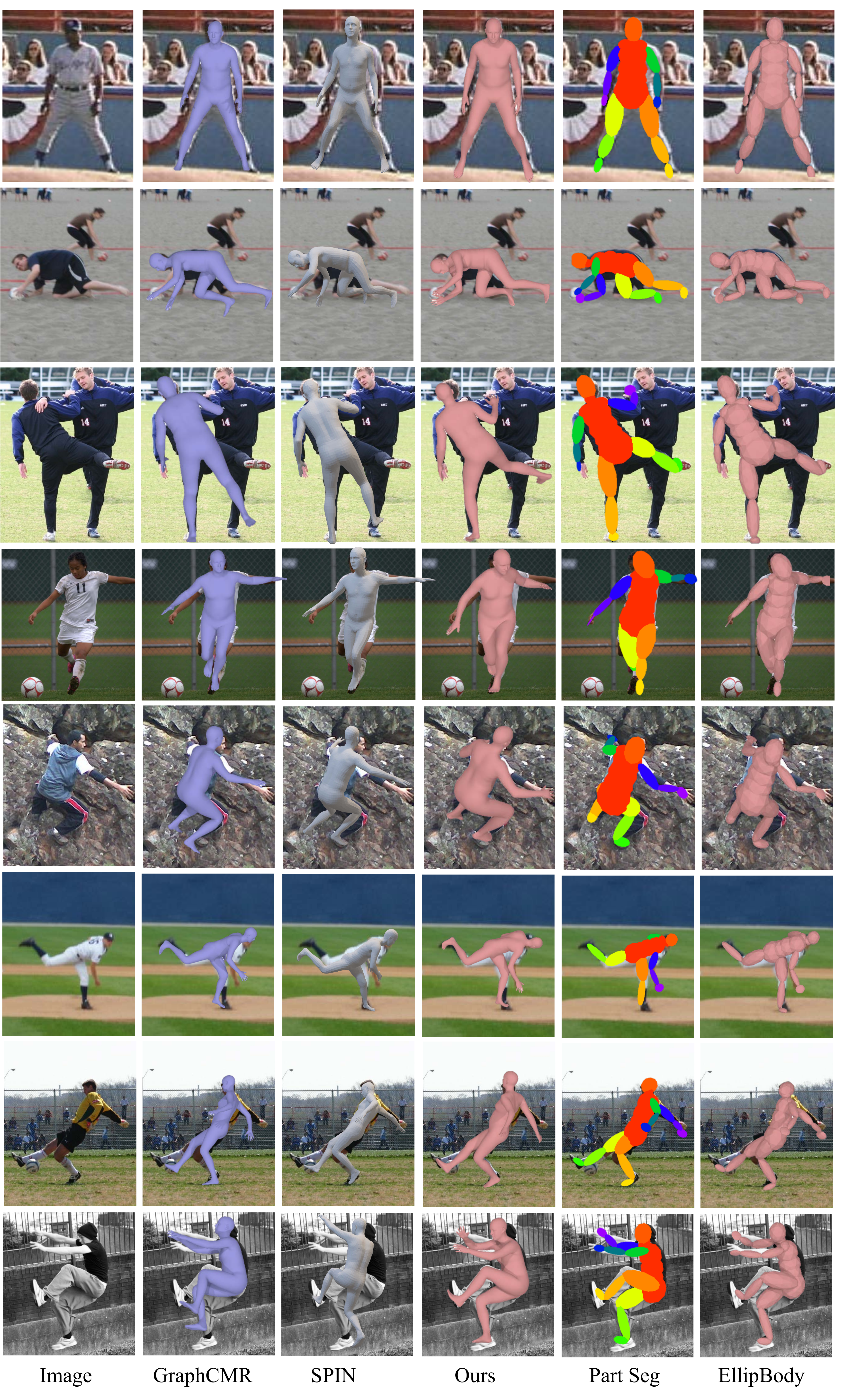}
	\end{center}
	\caption{\textbf{Qualitative Results.} Comparison with recent approaches for LSP datasets. Images from left to right indicate: (1) original image, (2) GraphCMR \cite{kolotouros2019convolutional}, (3) SPIN\cite{kolotouros2019learning} (4) registered SMPL on EllipBody, (4) part segmentation of EllipBody, (5) EllipBody. }
	\vspace{-3mm}
	\label{fig:qualitative}
\end{figure*}

%% file: main.bbl
\newcommand{\etalchar}[1]{$^{#1}$}
\begin{thebibliography}{\uppercase{DGLVG14}}

\bibitem[AARS13]{amin2013multi}
\textsc{Amin S., Andriluka M., Rohrbach M., Schiele B.}:
\newblock Multi-view pictorial structures for 3d human pose estimation.
\newblock In \emph{BMVC} (2013), vol.~1.

\bibitem[AB15]{akhter2015pose}
\textsc{Akhter I., Black M.~J.}:
\newblock Pose-conditioned joint angle limits for 3d human pose reconstruction.
\newblock In \emph{Proceedings of the IEEE conference on computer vision and
  pattern recognition} (2015), pp.~1446--1455.

\bibitem[APGS14]{andriluka20142d}
\textsc{Andriluka M., Pishchulin L., Gehler P., Schiele B.}:
\newblock 2d human pose estimation: New benchmark and state of the art
  analysis.
\newblock In \emph{Proceedings of the IEEE Conference on computer Vision and
  Pattern Recognition} (2014), pp.~3686--3693.

\bibitem[ASK{\etalchar{*}}05]{anguelov2005scape}
\textsc{Anguelov D., Srinivasan P., Koller D., Thrun S., Rodgers J., Davis J.}:
\newblock Scape: shape completion and animation of people.
\newblock In \emph{ACM Transactions on Graphics (TOG)} (2005), ACM.

\bibitem[BAA{\etalchar{*}}14]{belagiannis20143d}
\textsc{Belagiannis V., Amin S., Andriluka M., Schiele B., Navab N., Ilic S.}:
\newblock 3d pictorial structures for multiple human pose estimation.
\newblock In \emph{Proceedings of the IEEE Conference on Computer Vision and
  Pattern Recognition} (2014), pp.~1669--1676.

\bibitem[BKL{\etalchar{*}}16]{bogo2016keep}
\textsc{Bogo F., Kanazawa A., Lassner C., Gehler P., Romero J., Black M.~J.}:
\newblock Keep it smpl: Automatic estimation of 3d human pose and shape from a
  single image.
\newblock In \emph{European Conference on Computer Vision} (2016), Springer,
  pp.~561--578.

\bibitem[BM92]{besl1992method}
\textsc{Besl P.~J., McKay N.~D.}:
\newblock Method for registration of 3-d shapes.
\newblock In \emph{Sensor fusion IV: control paradigms and data structures}
  (1992), vol.~1611, International Society for Optics and Photonics,
  pp.~586--606.

\bibitem[BSC13]{burenius20133d}
\textsc{Burenius M., Sullivan J., Carlsson S.}:
\newblock 3d pictorial structures for multiple view articulated pose
  estimation.
\newblock In \emph{Proceedings of the IEEE Conference on Computer Vision and
  Pattern Recognition} (2013), pp.~3618--3625.

\bibitem[CSWS17]{cao2017realtime}
\textsc{Cao Z., Simon T., Wei S.-E., Sheikh Y.}:
\newblock Realtime multi-person 2d pose estimation using part affinity fields.
\newblock In \emph{Proceedings of the IEEE conference on computer vision and
  pattern recognition} (2017), pp.~7291--7299.

\bibitem[DGKS76]{daniel1976reorthogonalization}
\textsc{Daniel J.~W., Gragg W.~B., Kaufman L., Stewart G.~W.}:
\newblock Reorthogonalization and stable algorithms for updating the
  gram-schmidt qr factorization.
\newblock \emph{Mathematics of Computation 30}, 136 (1976), 772--795.

\bibitem[DGLVG14]{dantone2014body}
\textsc{Dantone M., Gall J., Leistner C., Van~Gool L.}:
\newblock Body parts dependent joint regressors for human pose estimation in
  still images.
\newblock \emph{IEEE Transactions on pattern analysis and machine intelligence
  36}, 11 (2014), 2131--2143.

\bibitem[dLGPF08]{de2008model}
\textsc{de~La~Gorce M., Paragios N., Fleet D.~J.}:
\newblock Model-based hand tracking with texture, shading and self-occlusions.
\newblock In \emph{2008 IEEE Conference on Computer Vision and Pattern
  Recognition} (2008), IEEE, pp.~1--8.

\bibitem[FH05]{felzenszwalb2005pictorial}
\textsc{Felzenszwalb P.~F., Huttenlocher D.~P.}:
\newblock Pictorial structures for object recognition.
\newblock \emph{International journal of computer vision 61}, 1 (2005), 55--79.

\bibitem[GWBB09]{guan2009estimating}
\textsc{Guan P., Weiss A., Balan A.~O., Black M.~J.}:
\newblock Estimating human shape and pose from a single image.
\newblock In \emph{2009 IEEE 12th International Conference on Computer Vision}
  (2009), IEEE, pp.~1381--1388.

\bibitem[HXM{\etalchar{*}}19]{habibie2019wild}
\textsc{Habibie I., Xu W., Mehta D., Pons-Moll G., Theobalt C.}:
\newblock In the wild human pose estimation using explicit 2d features and
  intermediate 3d representations.
\newblock In \emph{Proceedings of the IEEE Conference on Computer Vision and
  Pattern Recognition} (2019), pp.~10905--10914.

\bibitem[HZRS16]{he2016deep}
\textsc{He K., Zhang X., Ren S., Sun J.}:
\newblock Deep residual learning for image recognition.
\newblock In \emph{Proceedings of the IEEE conference on computer vision and
  pattern recognition} (2016), pp.~770--778.

\bibitem[IPOS13]{ionescu2014human3}
\textsc{Ionescu C., Papava D., Olaru V., Sminchisescu C.}:
\newblock Human3. 6m: Large scale datasets and predictive methods for 3d human
  sensing in natural environments.
\newblock \emph{IEEE transactions on pattern analysis and machine intelligence
  36}, 7 (2013), 1325--1339.

\bibitem[JE10]{Johnson10}
\textsc{Johnson S., Everingham M.}:
\newblock Clustered pose and nonlinear appearance models for human pose
  estimation.
\newblock In \emph{BMVC} (2010).
\newblock doi:10.5244/C.24.12.

\bibitem[KBJM18]{Kanazawa_2018_CVPR}
\textsc{Kanazawa A., Black M.~J., Jacobs D.~W., Malik J.}:
\newblock End-to-end recovery of human shape and pose.
\newblock In \emph{Proceedings of the IEEE Conference on Computer Vision and
  Pattern Recognition} (2018), pp.~7122--7131.

\bibitem[KPBD19]{kolotouros2019learning}
\textsc{Kolotouros N., Pavlakos G., Black M.~J., Daniilidis K.}:
\newblock Learning to reconstruct 3d human pose and shape via model-fitting in
  the loop.
\newblock In \emph{Proceedings of the IEEE International Conference on Computer
  Vision} (2019), pp.~2252--2261.

\bibitem[KPD19]{kolotouros2019convolutional}
\textsc{Kolotouros N., Pavlakos G., Daniilidis K.}:
\newblock Convolutional mesh regression for single-image human shape
  reconstruction.
\newblock In \emph{Proceedings of the IEEE Conference on Computer Vision and
  Pattern Recognition} (2019), pp.~4501--4510.

\bibitem[KUH18]{kato2018renderer}
\textsc{Kato H., Ushiku Y., Harada T.}:
\newblock Neural 3d mesh renderer.
\newblock In \emph{Proceedings of the IEEE Conference on Computer Vision and
  Pattern Recognition} (2018), pp.~3907--3916.

\bibitem[LB14]{loper2014opendr}
\textsc{Loper M.~M., Black M.~J.}:
\newblock Opendr: An approximate differentiable renderer.
\newblock In \emph{European Conference on Computer Vision} (2014), Springer,
  pp.~154--169.

\bibitem[LFL93]{liu1993kinematic}
\textsc{Liu K., Fitzgerald J.~M., Lewis F.~L.}:
\newblock Kinematic analysis of a stewart platform manipulator.
\newblock \emph{IEEE Transactions on Industrial Electronics 40}, 2 (1993),
  282--293.

\bibitem[LLCL19]{liu2019soft}
\textsc{Liu S., Li T., Chen W., Li H.}:
\newblock Soft rasterizer: A differentiable renderer for image-based 3d
  reasoning.
\newblock In \emph{Proceedings of the IEEE International Conference on Computer
  Vision} (2019), pp.~7708--7717.

\bibitem[LMB{\etalchar{*}}14]{lin2014microsoft}
\textsc{Lin T.-Y., Maire M., Belongie S., Hays J., Perona P., Ramanan D.,
  Doll{\'a}r P., Zitnick C.~L.}:
\newblock Microsoft coco: Common objects in context.
\newblock In \emph{European conference on computer vision} (2014), Springer,
  pp.~740--755.

\bibitem[LMR{\etalchar{*}}15]{loper2015smpl}
\textsc{Loper M., Mahmood N., Romero J., Pons-Moll G., Black M.~J.}:
\newblock Smpl: A skinned multi-person linear model.
\newblock \emph{ACM Transactions on Graphics (TOG) 34}, 6 (2015), 248.

\bibitem[LMSR17]{lin2017refinenet}
\textsc{Lin G., Milan A., Shen C., Reid I.}:
\newblock Refinenet: Multi-path refinement networks for high-resolution
  semantic segmentation.
\newblock In \emph{Proceedings of the IEEE conference on computer vision and
  pattern recognition} (2017), pp.~1925--1934.

\bibitem[LRK{\etalchar{*}}17]{lassner2017unite}
\textsc{Lassner C., Romero J., Kiefel M., Bogo F., Black M.~J., Gehler P.~V.}:
\newblock Unite the people: Closing the loop between 3d and 2d human
  representations.
\newblock In \emph{Proceedings of the IEEE conference on computer vision and
  pattern recognition} (2017), pp.~6050--6059.

\bibitem[LS88]{lee1988kinematic}
\textsc{Lee K.-M., Shah D.~K.}:
\newblock Kinematic analysis of a three-degrees-of-freedom in-parallel actuated
  manipulator.
\newblock \emph{IEEE Journal on Robotics and Automation 4}, 3 (1988), 354--360.

\bibitem[MHRL17]{martinez_2017_3dbaseline}
\textsc{Martinez J., Hossain R., Romero J., Little J.~J.}:
\newblock A simple yet effective baseline for 3d human pose estimation.
\newblock In \emph{Proceedings of the IEEE International Conference on Computer
  Vision} (2017), pp.~2640--2649.

\bibitem[NYD16]{newell2016stacked}
\textsc{Newell A., Yang K., Deng J.}:
\newblock Stacked hourglass networks for human pose estimation.
\newblock In \emph{European Conference on Computer Vision} (2016), Springer,
  pp.~483--499.

\bibitem[OLPM{\etalchar{*}}18]{omran2018neural}
\textsc{Omran M., Lassner C., Pons-Moll G., Gehler P., Schiele B.}:
\newblock Neural body fitting: Unifying deep learning and model based human
  pose and shape estimation.
\newblock In \emph{2018 international conference on 3D vision (3DV)} (2018),
  IEEE, pp.~484--494.

\bibitem[PCG{\etalchar{*}}19]{SMPL-X:2019}
\textsc{Pavlakos G., Choutas V., Ghorbani N., Bolkart T., Osman A.~A., Tzionas
  D., Black M.~J.}:
\newblock Expressive body capture: 3d hands, face, and body from a single
  image.
\newblock In \emph{Proceedings of the IEEE Conference on Computer Vision and
  Pattern Recognition} (2019), pp.~10975--10985.

\bibitem[PZD18]{pavlakos2018ordinal}
\textsc{Pavlakos G., Zhou X., Daniilidis K.}:
\newblock Ordinal depth supervision for 3d human pose estimation.
\newblock In \emph{Proceedings of the IEEE Conference on Computer Vision and
  Pattern Recognition} (2018), pp.~7307--7316.

\bibitem[PZDD17]{pavlakos2017coarse}
\textsc{Pavlakos G., Zhou X., Derpanis K.~G., Daniilidis K.}:
\newblock Coarse-to-fine volumetric prediction for single-image 3d human pose.
\newblock In \emph{Proceedings of the IEEE Conference on Computer Vision and
  Pattern Recognition} (2017), pp.~7025--7034.

\bibitem[PZZD18a]{pavlakos2018learning}
\textsc{Pavlakos G., Zhu L., Zhou X., Daniilidis K.}:
\newblock Learning to estimate 3d human pose and shape from a single color
  image.
\newblock In \emph{Proceedings of the IEEE Conference on Computer Vision and
  Pattern Recognition} (2018), pp.~459--468.

\bibitem[PZZD18b]{Pavlakos_2018_CVPR}
\textsc{Pavlakos G., Zhu L., Zhou X., Daniilidis K.}:
\newblock Learning to estimate 3d human pose and shape from a single color
  image.
\newblock In \emph{Proceedings of the IEEE Conference on Computer Vision and
  Pattern Recognition} (2018), pp.~459--468.

\bibitem[RKS12]{ramakrishna2012reconstructing}
\textsc{Ramakrishna V., Kanade T., Sheikh Y.}:
\newblock Reconstructing 3d human pose from 2d image landmarks.
\newblock In \emph{European conference on computer vision} (2012), Springer,
  pp.~573--586.

\bibitem[SHRB11]{straka2011skeletal}
\textsc{Straka M., Hauswiesner S., R{\"u}ther M., Bischof H.}:
\newblock Skeletal graph based human pose estimation in real-time.
\newblock In \emph{BMVC} (2011), pp.~1--12.

\bibitem[SIHB12]{sigal2012loose}
\textsc{Sigal L., Isard M., Haussecker H., Black M.~J.}:
\newblock Loose-limbed people: Estimating 3d human pose and motion using
  non-parametric belief propagation.
\newblock \emph{International journal of computer vision 98}, 1 (2012), 15--48.

\bibitem[SSLW17]{sun2017compositional}
\textsc{Sun X., Shang J., Liang S., Wei Y.}:
\newblock Compositional human pose regression.
\newblock In \emph{Proceedings of the IEEE International Conference on Computer
  Vision} (2017), pp.~2602--2611.

\bibitem[SXLW19]{sun2019deep}
\textsc{Sun K., Xiao B., Liu D., Wang J.}:
\newblock Deep high-resolution representation learning for human pose
  estimation.
\newblock In \emph{Proceedings of the IEEE conference on computer vision and
  pattern recognition} (2019), pp.~5693--5703.

\bibitem[SXW{\etalchar{*}}18]{sun2018integral}
\textsc{Sun X., Xiao B., Wei F., Liang S., Wei Y.}:
\newblock Integral human pose regression.
\newblock In \emph{Proceedings of the European Conference on Computer Vision
  (ECCV)} (2018), pp.~529--545.

\bibitem[VCR{\etalchar{*}}18]{varol2018bodynet}
\textsc{Varol G., Ceylan D., Russell B., Yang J., Yumer E., Laptev I., Schmid
  C.}:
\newblock Bodynet: Volumetric inference of 3d human body shapes.
\newblock In \emph{Proceedings of the European Conference on Computer Vision
  (ECCV)} (2018), pp.~20--36.

\bibitem[WCL{\etalchar{*}}18]{wang2018drpose3d}
\textsc{Wang M., Chen X., Liu W., Qian C., Lin L., Ma L.}:
\newblock Drpose3d: depth ranking in 3d human pose estimation.
\newblock In \emph{Proceedings of the 27th International Joint Conference on
  Artificial Intelligence} (2018), pp.~978--984.

\bibitem[XJS19]{xiang2019monocular}
\textsc{Xiang D., Joo H., Sheikh Y.}:
\newblock Monocular total capture: Posing face, body, and hands in the wild.
\newblock In \emph{Proceedings of the IEEE Conference on Computer Vision and
  Pattern Recognition} (2019), pp.~10965--10974.

\bibitem[XWW18]{xiao2018simple}
\textsc{Xiao B., Wu H., Wei Y.}:
\newblock Simple baselines for human pose estimation and tracking.
\newblock In \emph{Proceedings of the European conference on computer vision
  (ECCV)} (2018), pp.~466--481.

\bibitem[XZT19]{xu2019denserac}
\textsc{Xu Y., Zhu S.-C., Tung T.}:
\newblock Denserac: Joint 3d pose and shape estimation by dense
  render-and-compare.
\newblock In \emph{Proceedings of the IEEE International Conference on Computer
  Vision} (2019), pp.~7760--7770.

\bibitem[ZB15]{zuffi2015stitched}
\textsc{Zuffi S., Black M.~J.}:
\newblock The stitched puppet: A graphical model of 3d human shape and pose.
\newblock In \emph{Proceedings of the IEEE Conference on Computer Vision and
  Pattern Recognition} (2015), pp.~3537--3546.

\bibitem[ZHS{\etalchar{*}}17]{zhou2017towards}
\textsc{Zhou X., Huang Q., Sun X., Xue X., Wei Y.}:
\newblock Towards 3d human pose estimation in the wild: a weakly-supervised
  approach.
\newblock In \emph{Proceedings of the IEEE International Conference on Computer
  Vision} (2017), pp.~398--407.

\bibitem[ZZLD16]{zhou2017sparse}
\textsc{Zhou X., Zhu M., Leonardos S., Daniilidis K.}:
\newblock Sparse representation for 3d shape estimation: A convex relaxation
  approach.
\newblock \emph{IEEE transactions on pattern analysis and machine intelligence
  39}, 8 (2016), 1648--1661.

\end{thebibliography}
